\documentclass[11pt]{article}

\usepackage[final]{acl}

\usepackage{times}
\usepackage{latexsym}
\usepackage{float}

\usepackage[T1]{fontenc}

\usepackage[utf8]{inputenc}

\usepackage{microtype}

\usepackage{inconsolata}

\usepackage{graphicx}
\usepackage{subcaption}
\usepackage{amsmath}
\usepackage{amssymb}
\usepackage{algorithm}
\usepackage{algorithmic}
\usepackage{tabularx}
\usepackage{booktabs} 
\usepackage{pifont} 
\usepackage{multirow} 
\usepackage{enumitem}
\newcommand{\cmark}{\ding{51}}
\newcommand{\xmark}{\ding{55}}

\usepackage{xcolor}

\title{LogosKG: Hardware-Optimized Scalable and Interpretable Knowledge Graph Retrieval}

\author{
  \textbf{He Cheng\textsuperscript{1}},
  \textbf{Yifu Wu\textsuperscript{1}},
  \textbf{Saksham Khatwani\textsuperscript{1,2}},
  \textbf{Maya Kruse\textsuperscript{1}},
  \textbf{Dmitriy Dligach\textsuperscript{3}}, \\
  \textbf{Timothy A. Miller\textsuperscript{4,5}},
  \textbf{Majid Afshar\textsuperscript{6}},
  \textbf{Yanjun Gao\textsuperscript{1}\thanks{Corresponding author}} \\
  \textsuperscript{1}LARK Lab, University of Colorado Anschutz, 
  \textsuperscript{2}University of Colorado Boulder, \\
  \textsuperscript{3}Loyola University Chicago, 
  \textsuperscript{4}Harvard Medical School,\\
  \textsuperscript{5}Boston Children's Hospital, 
  \textsuperscript{6}University of Wisconsin-Madison \\
  \texttt{\{he.2.cheng, yanjun.gao\}@cuanschutz.edu}
}

\begin{document}
\maketitle
\begin{abstract}
Knowledge graphs (KGs) are increasingly integrated with large language models (LLMs) to provide structured, verifiable reasoning. A core operation in this integration is multi-hop retrieval, yet existing systems struggle to balance efficiency, scalability, and interpretability. We introduce \textsc{LogosKG}, a novel, hardware-aligned framework that enables scalable and interpretable $k$-hop retrieval on large KGs by building on symbolic KG formulations and executing traversal as hardware-efficient operations over decomposed subject, object, and relation representations. To scale to billion-edge graphs, \textsc{LogosKG} integrates degree-aware partitioning, cross-graph routing, and on-demand caching. Experiments show substantial efficiency gains over CPU and GPU baselines without loss of retrieval fidelity. With proven performance in KG retrieval, a downstream two-round KG-LLM interaction demonstrates how \textsc{LogosKG} enables large-scale, evidence-grounded analysis of how KG topology, such as hop distribution and connectivity, shapes the alignment between structured biomedical knowledge and LLM diagnostic reasoning, thereby opening the door for next-generation KG-LLM integration. The source code is publicly available at \url{https://github.com/LARK-NLP-Lab/LogosKG}, and an online demo is available at \url{https://lark-nlp-lab-logoskg.hf.space/}.

\end{abstract}

\section{Introduction}
For decades, knowledge graphs (KGs) have served as a foundation for structured knowledge representation, linking concepts through relations across domains such as social networks \citep{cai2023knowledge}, biomedicine \citep{lu2025biomedical}, and recommendation systems \citep{wang-etal-2025-knowledge-graph}. With the rise of large language models (LLMs), KGs have gained renewed importance as an external symbolic knowledge source that complements LLMs' statistical reasoning in tasks like retrieval-augmented generation (RAG) \citep{liu-etal-2025-hoprag, sharma2024og}, knowledge verification \citep{pham-etal-2025-verify, dammu2024claimver}, and reasoning \citep{wang2025reasoning, wu2025kg}. In high-stakes domains such as medical diagnosis, where LLM reliability directly affects patient safety, KGs are increasingly used to ground model predictions in verified biomedical knowledge \citep{jia2024medikal, zuo2025kg4diagnosis, rezaei2025adaptive}.

\begin{table}[t]
\centering
\scriptsize
\setlength{\tabcolsep}{1pt}
\resizebox{\linewidth}{!}{
\begin{tabular}{lccccc}
\toprule \toprule
\textbf{Method (year)} & \textbf{Matrix-} & \textbf{Scala} & \textbf{Path} & \textbf{Device} \\
& \textbf{based} & \textbf{-bility}& \textbf{Reconstruction} & \\ 
\midrule
Neo4j \citep{neo4j}              & \xmark & \cmark                 & \cmark & CPU       \\
TigerGraph \citep{deutsch2019tigergraph}          & \xmark & \cmark                 & \cmark & CPU       \\ \midrule
GraphBLAS \citep{graphblas-hpec}           & \cmark & \xmark                 & \xmark & CPU       \\ \midrule
igraph \citep{csardi2006igraph}              & \xmark & \xmark                 & \cmark & CPU       \\
NetworkX \citep{hagberg2008exploring}           & \xmark & \xmark                 & \cmark & CPU       \\
graph-tool \citep{peixoto_graph-tool_2014}          & \xmark & \xmark                 & \cmark & CPU       \\
SNAP \citep{leskovec2016snap}               & \xmark & \cmark                 & \cmark & CPU       \\ \midrule
cuGraph \citep{cugraph}            & \cmark & \xmark                 & \cmark & GPU       \\
DGL \citep{wang2019deep}               & \xmark & \cmark                 & \cmark & GPU       \\
PyG  \cite{fey2019fast}               & \xmark & \cmark                 & \xmark & GPU       \\
\midrule
\textbf{LogosKG (ours, 2026)} & \cmark & \cmark & \cmark & CPU/GPU \\
\bottomrule \bottomrule
\end{tabular}
}
\vspace{-.1in}

\caption{\small Comparison of retrieval systems and libraries. Matrix-based indicates the use of linear-algebra primitives for retrieval. \textit{Note:} Entries reflect the default design under a single-machine setting. Scalability indicates the ability to process graphs that exceed single-machine memory limits. Unavailable features can be achieved with additional processing, at the cost of time and memory.}
\label{tab: retrieval-comparison}
\end{table}

\begin{figure*}[htbp]
    \centering
    \includegraphics[width=0.98\linewidth]{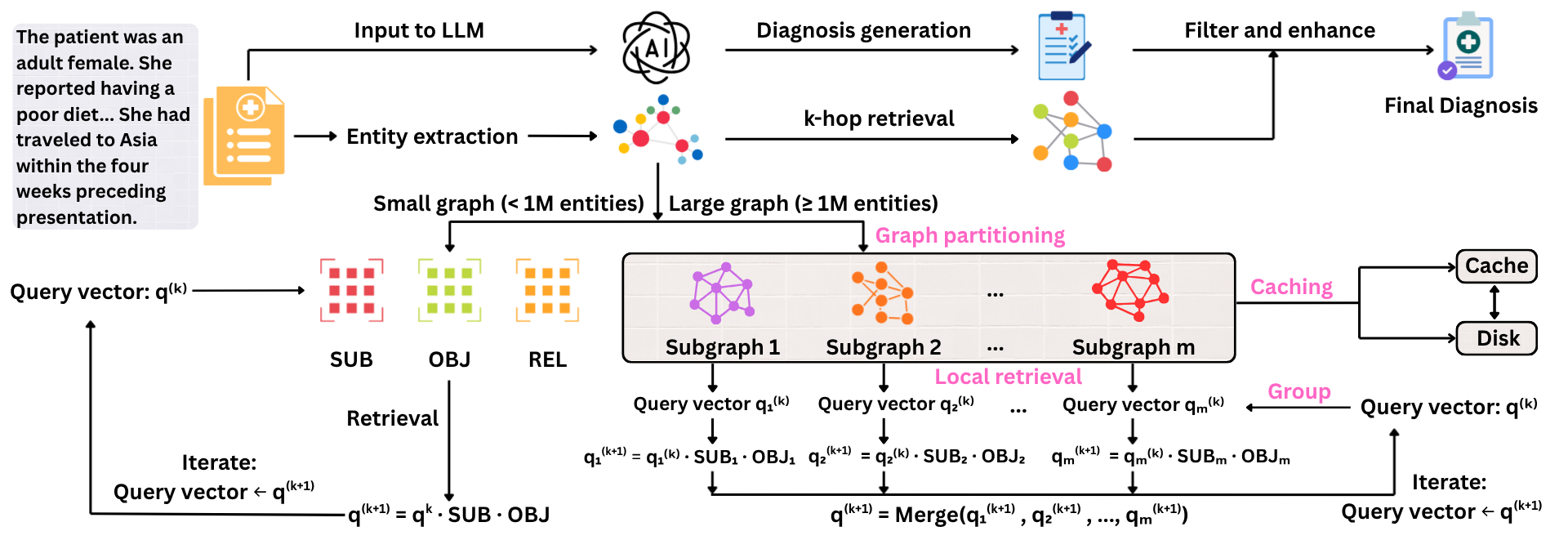}
    \vspace{-.1in}
    \caption{
        \small Overview of the LogosKG retrieval framework. LogosKG adopts a linear algebra–based retrieval method that supports both small and large KGs. For large graphs, LogosKG incorporates graph partitioning, cross-graph routing, and on-demand caching to enable efficient multi-hop retrieval across distributed subgraphs. The retrieved evidence can provide knowledge-grounded support for medical diagnosis generation. }
    \label{fig:logoskg}
\end{figure*}

A fundamental operation of KGs is \emph{multi-hop retrieval}, which connects distant concepts through intermediate entities and relations. Traditional graph traversal algorithms such as depth-first search (DFS) and breadth-first search (BFS) can handle small graphs but quickly become infeasible as graph size grows, with cost scaling as $O(|V| + |E|)$ and reachable entities growing exponentially with hop depth. 
\textit{Memory} is another major bottleneck: for instance, biomedical KGs used in our experiments (\textbf{UMLS}~\cite{bodenreider2004unified}, 407K nodes, 3.4M edges; and the $100\times$ larger \textbf{PubMedKG}~\cite{xu2020building,xu2025pubmed}, 54.4M nodes, 86.5M edges) occupy 1.5 GB and 23.5 GB of memory even before traversal.
A two-hop expansion from a high-degree concept in UMLS (on average $\approx$33k 1-hop neighbors) can involve over $10^9$ reachable edges, consuming tens of gigabytes of memory to materialize adjacency information. Such exponential growth in traversal and storage defines the \textit{core systems challenge} in scaling multi-hop retrieval, forcing prior work to operate on limited subgraphs~\cite{gao2025leveraging,chang2020benchmark}.

Numerous systems have been developed to improve retrieval efficiency (Table~\ref{tab: retrieval-comparison}) emphasizing three dimensions: (1) \textit{matrix-based} graph representation replacing pointer-based data structures with matrix or tensor operations for vectorized computation that maps naturally onto parallel hardware such as multi-core CPUs or GPUs (e.g., GraphBLAS-based systems); (2) \emph{scalability}, supported by mechanisms such as graph partitioning to handle billion-edge graphs (e.g., DGL, PyG); and (3) \emph{path reconstruction} enabling interpretable reasoning by recovering intermediate entities and relations (e.g., Neo4j, TigerGraph). However, most systems optimize only one or two of these aspects and often rely on distributed infrastructures rather than single-device efficiency.

\textbf{Our Approach.} To achieve scalable and interpretable KG retrieval, we present \textbf{\textsc{LogosKG}}, a \textit{novel} hardware-aligned framework that rethinks graph traversal at the system level (as shown in Figure~\ref{fig:logoskg}). While symbolic graph reasoning has been explored in prior work~\cite{Cohen2020Scalable}, \textsc{LogosKG} extends this methodology into a unified, end-to-end system that makes high-hop traversal practical on large, densely connected KGs using single-device hardware, allowing contiguous data layouts and kernel-level parallelism on CPUs and GPUs. 
For large graphs, \textsc{LogosKG} implements degree-aware partitioning and cross-graph routing guided by memory-locality principles, while on-demand caching further reduces I/O overhead, achieving $\mathcal{O}(|\mathcal{E}|\log|\mathcal{E}| + |\mathcal{T}|)$ complexity 
(where $\mathcal{E}$ is the set of entities and $\mathcal{T}$ is the set of triples). 
Finally, \textsc{LogosKG} stores intermediate entities and relations that enable path reconstruction. Together, these innovations transform a theoretical formulation into a practical and scalable framework for large-scale KG retrieval. 

We evaluate \textsc{LogosKG} against existing libraries and systems (Table~\ref{tab: retrieval-comparison}) across retrieval efficiency, scalability, and interpretability.
Experiments focus on large, semantically rich biomedical KGs, which present realistic challenges for dense connectivity, heterogeneous relations, and interpretable reasoning.
Although the experiments use biomedical data, \textsc{LogosKG} is domain-agnostic and applicable to any large structured graph. 

\textsc{LogosKG} removes a fundamental system bottleneck in scalable high-hop retrieval, thereby enabling the study of LLM reasoning over high-hop KGs, independent of whether downstream selection or reasoning components are learned or nonlearned. We instantiate this capacity through a systematic study of the KG-LLM setup for clinical diagnosis prediction, using a two-round interaction setup to examine how LLM responds to KG structure under increasing hop depths. We summarize the main contributions of this paper as follows: 


\noindent 
\begin{itemize}[leftmargin=*, nosep] 
\item \textbf{Scalable system capability for large KG traversal}: We present \textsc{LogosKG}, a hardware-aligned framework that enables deterministic, interpretable high-hop retrieval at scale on single-device hardware (\S\ref{sec:logoskg}).
\item \textbf{Comprehensive system evaluation:} We benchmark retrieval fidelity, efficiency, and scalability across CPU and GPU baselines (\S\ref{sec:logoskg_results}).
\item \textbf{Analysis of KG-LLM interaction under high-hop regimes:} Leveraging \textsc{LogosKG}, we perform a systematic study of how LLM predictions interact with deep KG structures through a two-round interaction investigation (\S\ref{sec:interaction_regimes}). 
\end{itemize}

\textsc{LogosKG} serves as a general systems backbone for high-hop KG retrieval and can support both learned and non-learned refinement and reasoning strategies at scale. While KG-LLM interaction is a complex research question in its own right, we include analyses of both non-learned and learned high-hop KG-LLM designs in the Appendix (\S\ref{appendix:llm_round2_sft}–\ref{appendix:logoskg_sft}), illustrating that \textsc{LogosKG} makes such studies feasible. 



\section{Related Work}

Existing systems and libraries can be classified by their architecture, ranging from database-backed engines and computation-focused libraries to graph analysis tools and GPU-based frameworks. \textbf{Database-backed engines} such as Neo4j and TigerGraph \citep{deutsch2019tigergraph} provide expressive query languages but require significant infrastructure for sharding and incur runtime overhead from query parsing pipelines and transaction logs. \textbf{Computation-focused libraries} such as GraphBLAS \citep{graphblas-hpec} use sparse matrix operations to make multi-hop retrieval efficient, yet lack native support for full path reconstruction due to the loss of edge provenance during aggregation. \textbf{Graph analysis tools}, such as igraph \citep{csardi2006igraph}, NetworkX \citep{hagberg2008exploring}, and SNAP \citep{leskovec2016snap}, offer flexible APIs but rely on memory-bound pointer-chasing algorithms that do not scale effectively. Finally, \textbf{GPU-based frameworks} such as cuGraph, DGL \citep{wang2019deep}, and PyG \citep{fey2019fast} enable fast single-machine computation but prioritize dense tensor-based training over retrieval; their reliance on classical search backends limits efficiency for multi-hop reasoning.

\subsection{KGs as Verifiers for LLMs}
LLMs are prone to hallucinations and unverified claims, motivating research on KG-based verification and fact-checking. Recent methods leverage KGs to ground textual claims in structured evidence and construct multi-hop reasoning chains that link claims to supporting facts. Examples include FactKG \citep{kim2023factkg} and GraphCheck \citep{chen2025graphcheck} for long-context fact-checking, ClaimVer \citep{dammu2024claimver} and Verify-in-the-Graph \citep{pham2025verify} for explainable evidence attribution and entity disambiguation, and GraphFC \citep{huang2025graph} and FactCheck \citep{shami2025fact} for integrating graph reasoning with LLM-based verification.
In contrast, our work exposes the raw topology of KGs and reveals a structural gap between symbolic knowledge organization and neural reasoning.  



\section{\textsc{LogosKG}} 
\label{sec:logoskg}
\subsection{Problem Statement}
\noindent \textbf{Knowledge Graph.}
A \emph{KG} is defined as a directed multi-relational graph 
\[
\mathcal{G} = (\mathcal{E}, \mathcal{R}, \mathcal{T}),
\]
where $\mathcal{E}$ is the set of entities, $\mathcal{R}$ is the set of relation types, and 
$\mathcal{T} \subseteq \mathcal{E} \times \mathcal{R} \times \mathcal{E}$ 
is the set of observed triples. 
Each triple $(e_s, r, e_o) \in \mathcal{T}$ denotes that subject entity $e_s \in \mathcal{E}$ is linked to object entity $e_o \in \mathcal{E}$ through relation $r \in \mathcal{R}$.

\noindent \textbf{$k$-hop Retrieval.}
Given a query $\mathbf{q} \in \mathcal{E}$ consisting of a set of entities, the goal of $k$-hop retrieval is to identify all entities reachable from any $e_q \in \mathbf{q}$ within $k$ relational steps. In practice, querying large KGs can be time-consuming. The task is to retrieve $k$-hop entities in large KGs while keeping query latency as low as possible.

\subsection{KGs Decomposition}
\label{section: single-graph retrieval}
To support efficient $k$-hop retrieval, a KG could be represented using three sparse incidence matrices encoding subjects, objects, and relations \citep{Cohen2020Scalable}. This decomposition transforms heavy graph traversal into lightweight sparse matrix operations, as shown in Figure~\ref{fig:logoskg}.
\vspace{.02in}

\noindent \textbf{Subject matrix.}
The \emph{subject matrix} $\mathbf{SUB} \in \{0,1\}^{|\mathcal{E}|\times|\mathcal{T}|}$ is defined as:

{\small
\begin{equation}
[\mathbf{SUB}]_{i,t} =
\begin{cases}
1, & \text{if $e_i$ is the subject of $t$}, \\
0, & \text{otherwise}.
\end{cases}
\label{eq:sub}
\end{equation}
} 

where $e_i$ denotes an entity and $t$ a triplet, with rows as entities and columns as triplets. 

\vspace{.4em}

\noindent \textbf{Object matrix.}
The \emph{object matrix} $\mathbf{OBJ} \in \{0,1\}^{|\mathcal{T}|\times|\mathcal{E}|}$ is defined as:
{\small
\begin{equation}
[\mathbf{OBJ}]_{t,j} =
\begin{cases}
1, & \text{if $e_j$ is the object of $t$}, \\
0, & \text{otherwise}.
\end{cases}
\label{eq:obj}
\end{equation}
}

where $e_j$ is an entity and $t$ a triplet, with rows as triplets and columns as entities. 

\vspace{.4em}

\noindent \textbf{Relation matrix.}
The \emph{relation matrix} $\mathbf{REL} \in \{0,1\}^{|\mathcal{T}|\times|\mathcal{R}|}$ encodes the relation type of each triplet, and is defined as:

{\small 
\begin{equation}
[\mathbf{REL}]_{t,r} =
\begin{cases}
1, & \text{if $t$ uses relation $r$}, \\
0, & \text{otherwise}.
\end{cases}
\label{eq:rel}
\end{equation}
}
where $t$ denotes a triplet and $r$ a relation type. Each row corresponds to a triplet and each column to a relation, enabling recovery of relation paths in multi-hop retrieval.

\subsection{Efficient Retrieval}
\noindent \textbf{One-hop retrieval.}
Define query vector as $\mathbf{q}^{(0)} \in \{0,1\}^{1\times|\mathcal{E}|}$, where $\mathbf{q}^{(0)}[i]=1$ if entity $e_i$ is included in the query. Multiplying by $\mathbf{SUB}$ activates the triplets starting from these entities, and multiplying further by $\mathbf{OBJ}$ yields the reachable entities:

{\small 
\[
\begin{aligned}
\mathbf{q}^{(0)} \cdot \mathbf{SUB} &\;\mapsto\; \mathbf{t}^{(1)} (\text{active triplets}), \\
 \mathbf{t}^{(1)} \cdot \mathbf{OBJ} &\;\mapsto\; \mathbf{q}^{(1)} (\text{active entities}).
\end{aligned}
\]
}

Formally,
{\small 
\begin{equation}
\mathbf{q}^{(1)} = \mathbf{q}^{(0)} \cdot \mathbf{SUB} \cdot \mathbf{OBJ}.
\label{eq:hop}
\end{equation}
}

Repeating this operation $k$ times yields \textbf{$k$-hop retrieval}, denoted as: 
$\mathbf{q}^{(k)} = \mathbf{q}^{(0)} \cdot (\mathbf{SUB}\cdot\mathbf{OBJ})^k$, where $\mathbf{q}^{(k)} \in \{0,1\}^{1\times|\mathcal{E}|}$ and nonzero entries indicate the entities reachable exactly at $k$ steps.

\vspace{0.2em}
\noindent \textbf{Path reconstruction.}
At any hop $h$, we retrieve the activated triplets by

{\small 
\begin{equation}
\mathbf{t}^{(h)} = \mathbf{q}^{(h-1)} \cdot \mathbf{SUB}.
\label{eq:th}
\end{equation}
}

The set of nonzero indices is

{\small 
\begin{equation}
T_h = \{\, t \mid \mathbf{t}^{(h)}[t] = 1 \,\}.
\label{eq:Th}
\end{equation}
}

For each $t \in T_h$, we retrieve the subject entity, object entity, and relation of the triple directly from the incidence matrices:

{\small 
\begin{align}
s_h &= \{\, s \mid [\mathbf{SUB}]_{i,t}=1 \,\}, \nonumber \\
r_h &= \{\, r \mid [\mathbf{REL}]_{t,r}=1 \,\}, \nonumber \\
e_h &= \{\, o \mid [\mathbf{OBJ}]_{t,j}=1 \,\}.
\label{eq:components}
\end{align}
}

We build the path $s_h \xrightarrow{r_h} e_h$, and if $e_h$ also appears as a subject, we extend it at hop $h{+}1$. Repeating this yields complete paths of length $k$.

\begin{algorithm}[t!]
\caption{Cross-graph $k$-Hop Retrieval with On-Demand Caching}
\label{alg:cross_retrieval}
\small 
\begin{algorithmic}[1]
\REQUIRE Initial query vector $\mathbf{q}^{(0)}$, hop count $k$, subgraphs $\{\mathcal{G}_i\}_{i=1}^M$ with $(\mathbf{SUB}_i,\mathbf{OBJ}_i,\mathbf{REL}_i)$, metadata $P$, cache $C$
\ENSURE Query vector $\mathbf{q}^{(k)}$
\FOR{$\ell=0$ \TO $k-1$}
  \STATE Route nonzero entries of $\mathbf{q}^{(\ell)}$ to subgraphs via $P$
  \STATE Initialize $\mathbf{q}^{(\ell+1)} \leftarrow \mathbf{0}$
  \FORALL{subgraphs $\mathcal{G}_i$ with active entities}
    \STATE Load $\mathcal{G}_i$ into cache $C$ if absent (evict LRU if full)
    \STATE Restrict $\mathbf{q}^{(\ell)}$ to local subvector $\mathbf{q}^{(\ell)}_i$
    \STATE Compute local update $\mathbf{q}^{(\ell+1)}_i = \mathbf{q}^{(\ell)}_i \cdot \mathbf{SUB}_i \cdot \mathbf{OBJ}_i$
    \STATE Map $\mathbf{q}^{(\ell+1)}_i$ back to global indices
    \STATE $\mathbf{q}^{(\ell+1)} \leftarrow \mathbf{q}^{(\ell+1)} \lor \mathbf{q}^{(\ell+1)}_i$
  \ENDFOR
\ENDFOR
\RETURN $\mathbf{q}^{(k)}$
\end{algorithmic}
\end{algorithm}

\subsection{Scalable Retrieval}

Large-scale KGs can contain hundreds of millions or even billions of triplets, making it infeasible to load the entire graph into memory. To address this, we design a hierarchical system with degree-aware partitioning, cross-graph retrieval, and on-demand caching. Globally, queries are distributed across subgraphs; locally, retrieval within each subgraph uses sparse matrix operations. 

Specifically, a KG is partitioned into balanced subgraphs, which are processed independently, with a mapping table that tracks each entity’s subgraph assignment. During retrieval, queries are routed to their respective subgraphs for local multi-hop retrieval (\S\ref{section: single-graph retrieval}), and results are merged. On-demand caching loads only the required subgraphs into memory, keeping the rest on disk. This step is the key to efficient retrieval over billion-scale KG on limited hardware. 
 
\noindent \textbf{Degree-aware graph partitioning.}  
A KG can be partitioned into disjoint subgraphs $\{\mathcal{G}_i=(\mathcal{E}_i,\mathcal{R}_i,\mathcal{T}_i)\}_{i=1}^m$, where each $\mathcal{G}_i$ contains a subset of entities, relations, and triplets. Subject entities are assigned in a degree-aware manner so that high-degree subject entities are distributed evenly, avoiding bottlenecks caused by imbalance. To preserve KG structures, all triplets sharing the same subject entity are assigned to the same subgraph. The procedure runs in $O(|\mathcal{E}|\log|\mathcal{E}| + |\mathcal{T}|)$ time, including finding relations and sorting entities by degrees in $O(|\mathcal{E}|\log|\mathcal{E}| + |\mathcal{T}|)$ time and assigning triplets to subgraphs in $O(|\mathcal{T}|)$ time. We also maintain a metadata map $P:\mathcal{E}\to \{1,\ldots,m\}$ that records the partition assignment of each subject entity for efficient cross-partition retrieval. For each subgraph $\mathcal{G}_i$, we build local incidence matrices $\mathbf{SUB}_i$, $\mathbf{OBJ}_i$, and $\mathbf{REL}_i$ to support efficient retrieval within the partition. 

Our framework is not limited to this partitioning algorithm. In fact, LogosKG can support any partitioning strategy and integrate them with the following cross-graph routing and on-demand caching mechanisms, although retrieval latency may vary.


\noindent \textbf{Cross-subgraph retrieval.}    
Given a query entity vector $\mathbf{q}^{(0)}$, we first group entities by their assigned subgraphs using the metadata map $P$. For each subgraph $\mathcal{G}_i$, we build a local query vector $\mathbf{q}^{(0)}_i$ representing the subset of entities assigned to $\mathcal{G}_i$. Retrieval is then carried out independently within each subgraph using its local incidence matrices $\mathbf{SUB}_i$, $\mathbf{OBJ}_i$, and $\mathbf{REL}_i$. For any hop $k$, the update step follows the same principle as in the single-graph case:

{\small
\begin{equation}
\mathbf{t}^{(k)}_i = \mathbf{q}^{(k)}_i \cdot \mathbf{SUB}_i,
\end{equation}
\begin{equation}
\mathbf{q}^{(k+1)}_i = (\mathbf{q}^{(k)}_i \cdot \mathbf{SUB}_i) \cdot \mathbf{OBJ}_i.
\end{equation}
}
Entities $\mathbf{q}^{(k+1)}_i$ from all subgraphs are merged into a global query vector $\mathbf{q}^{(k+1)}$, then redistributed for the next hop. Repeating this $k$ times reconstructs the $k$-hop neighborhood while limiting computation to accessed subgraphs. 

\noindent \textbf{On-demand caching.}  
This mechanism makes sure subgraphs are loaded only when required. It avoids memory blow-up but introduces additional I/O overhead, since not all subgraphs needed for a query are guaranteed to be in memory.  

An in-memory cache of fixed capacity $n$ is implemented and capable of storing at most $n$ subgraphs, managed under a least-recently-used (LRU) policy. When a subgraph is requested, it is reused if present in the cache, or otherwise loaded from disk and inserted by evicting the least recently accessed subgraph. Each subgraph is stored on disk in a sparse format containing the incidence matrices. Cache hits require only sparse matrix multiplications, while cache misses incur the additional cost of disk I/O.  

Let $h$ denote the cache hit rate, $\tau_{\text{mm}}$ the time for in-memory sparse matrix multiplication, and $\tau_{\text{io}}$ the average disk load time. The expected retrieval cost per subgraph is

\vspace{-.07in}
{\small
\begin{equation}
\mathbb{E}[\tau_{\text{retrieval}}] 
= h \cdot \tau_{\text{mm}} + (1-h) \cdot (\tau_{\text{mm}} + \tau_{\text{io}}).
\end{equation}
}

We optimize batch processing for better cache efficiency by grouping queries with similar subgraph requirements. Each query’s subgraphs are identified via $P$, queries are reordered for joint processing, and results are restored afterward. This improves temporal locality, allowing subgraph reuse and reducing cache misses and I/O overhead. 

\noindent \textbf{Overall pipeline.}
The retrieval process begins with an initial query vector $\mathbf{q}^{(0)}$. At each hop $k$, we (i) map the active entities in $\mathbf{q}^{(h)}$ to their subgraphs using $P$, (ii) perform local retrieval within each subgraph using incidence matrices, (iii) merge the local results into the next global query vector $\mathbf{q}^{(h+1)}$, and (iv) repeat for $k$ hops. The procedure is summarized in Algorithm~\ref{alg:cross_retrieval}.

\noindent \textbf{Implementation Details} 
We implement \textsc{LogosKG} with three computational backends: Numba, SciPy, and Torch, where the Torch backend supports both CPU and GPU execution. Experiments were conducted on a HIPAA-compliant Linux server with Ubuntu 22.04 system, dual AMD EPYC 9454 48-Core CPUs (192 threads), 256 GB RAM, and two NVIDIA H100 NVL GPUs (94 GB VRAM each). 

\begin{table*}
\centering
\resizebox{\linewidth}{!}{
\begin{tabular}{@{}l*{5}{rr}@{}}
\toprule \toprule
\multirow{2}{*}{\textbf{Method}} &
\multicolumn{2}{c}{\textbf{Hop 1 (2000 ms)}} &
\multicolumn{2}{c}{\textbf{Hop 2 (4000 ms)}} &
\multicolumn{2}{c}{\textbf{Hop 3 (6000 ms)}} &
\multicolumn{2}{c}{\textbf{Hop 4 (8000 ms)}} &
\multicolumn{2}{c}{\textbf{Hop 5 (10000 ms)}} \\
\cmidrule(lr){2-3}\cmidrule(lr){4-5}\cmidrule(lr){6-7}\cmidrule(lr){8-9}\cmidrule(l){10-11}
& \textbf{QT (ms)} & \textbf{TR (\%)} & \textbf{QT (ms)} & \textbf{TR (\%)} & \textbf{QT (ms)} & \textbf{TR (\%)} & \textbf{QT (ms)} & \textbf{TR (\%)} & \textbf{QT (ms)} & \textbf{TR (\%)} \\
\midrule
\multicolumn{11}{l}{\emph{Baselines}} \\
\midrule
NetworkX          & \textbf{0.21}     & \textbf{0.00}  & \textbf{5.47}     & \textbf{0.00}  & 93.92    & 0.00  & 621.95   & 0.00  & 1511.28   & 0.00 \\
igraph            & \underline{1.15}  & \underline{0.00}  & 26.13    & 0.00  & 309.90   & 0.00  & 837.12   & 0.00  & 580.91    & 0.00 \\
graph-tool        & $>1458.79$ & 45.33 & $>1900.01$ & 10.00 & $>2141.41$ & 2.00 & $>2396.95$ & 0.67 & $>2306.34$ & 0.67 \\
SNAP              & 1.80     & 0.00  & \underline{10.02}    & \underline{0.00}  & 115.00   & 0.00  & 378.81   & 0.00  & 446.15    & 0.00 \\
GraphBLAS         & 3.03     & 0.00  & 43.64    & 0.00  & 291.89   & 0.00  & 528.07   & 0.00  & 415.43    & 0.00 \\
Neo4j             & $>923.86$ & 11.33 & $>1946.43$ & 20.00 & $>5739.02$ & 95.33 &  $>8000.00$           & 100      &   $>10000.00$          &  100     \\
cuGraph           & $>722.66$ & 3.33  & $>919.30$  & 0.67  & 1204.00  & 0.00  & 1504.82  & 0.00  & 1616.55   & 0.00 \\
DGL               & $>966.65$ & 10.00 & $>989.91$  & 0.67  & 1042.25  & 0.00  & 1121.70  & 0.00  & 1099.26   & 0.00 \\
PyG               & 249.66    & 0.00  & 271.46    & 0.00  & 365.90   & 0.00  & 646.96   & 0.00  & 735.34    & 0.00 \\
\midrule
\multicolumn{11}{l}{\emph{LogosKG family}} \\
\midrule
LogosKG (Numba)           & 12.28   & 0.00  & 28.72   & 0.00  & \underline{77.65}   & \underline{0.00}  & \underline{140.07}  & \underline{0.00}  & \underline{204.25}   & \underline{0.00} \\
LogosKG (SciPy)           & 13.81   & 0.00  & 34.97   & 0.00  & 104.21  & 0.00  & 289.61  & 0.00  & 677.23   & 0.00 \\
LogosKG (Torch-CPU)       & 526.55  & 0.00  & 884.89  & 0.00  & 1321.05 & 0.00  & 1803.18 & 0.00  & 2207.25  & 0.00 \\
LogosKG (Torch-GPU)       & 6.00    & 0.00  & 14.40   & 0.00  & \textbf{43.07}   & \textbf{0.00}  & \textbf{77.73}   & \textbf{0.00}  & \textbf{101.05}   & \textbf{0.00} \\
LogosKG (Large-Numba)     & 4.76    & 0.00  & 25.52   & 0.00  & 226.83  & 0.00  & 1411.72 & 0.00  & $>4085.72$ & 4.00 \\
LogosKG (Large-SciPy)     & 7.64    & 0.00  & 63.93   & 0.00  & 324.95  & 0.00  & $>1660.91$ & 0.67  & $>4532.75$ & 5.33 \\
LogosKG (Large-Torch-CPU) & 126.61  & 0.00  & $>1205.50$ & 0.67 & $>2551.68$ & 5.33 & $>4836.16$ & 22.00 & $>7467.56$ & 48.67 \\
LogosKG (Large-Torch-GPU) & 13.75   & 0.00  & 103.50  & 0.00  & 412.38  & 0.00  & 1634.93 & 0.00  & $>4482.14$ & 6.00 \\
\bottomrule \bottomrule
\end{tabular}
}
\caption{\small Retrieval efficiency comparison across hops. Timeout limits are fixed at 2000, 4000, 6000, 8000, and 10000 ms for 1–5 hops, respectively. All methods are run under the same CPU workload for fair comparison. For each hop, the best method is shown in bold and the second-best is underlined.}
\label{tab:kg_retrieval_h1to5}
\end{table*}

\begin{table}[ht!]
\centering
\small
\setlength{\tabcolsep}{8pt}
\resizebox{\columnwidth}{!}{
\begin{tabular}{lcccc}
\toprule \toprule
\textbf{Factor} & \textbf{Value} & \textbf{QT (ms)} & \textbf{Loads} & \textbf{Evicts} \\
\midrule
\multicolumn{5}{l}{\emph{Exp. 1: hops} (Numba, cache size $n=16$, batch size=50)} \\ \midrule
hops        & 1   & 3410.90   & 16   & 0    \\
hops        & 2   & 1610.78   & 16   & 0    \\
hops        & 3   & 6114.69   & 16   & 0    \\
hops        & 4   & 19592.08  & 16   & 0    \\
hops        & 5   & 62726.25  & 16   & 0    \\
\midrule
\multicolumn{5}{l}{\emph{Exp. 2: batch size} (Numba, cache size $n=16$, hops $k=2$)} \\ \midrule
batch size  & 1     & 100912.09 & 12   & 0    \\
batch size  & 10    & 5489.45   & 16   & 0    \\
batch size  & 25    & 1365.09   & 16   & 0    \\
batch size  & 50    & 1499.39   & 16   & 0    \\
batch size  & 100   & 1444.68   & 16   & 0    \\
batch size  & 150   & 1483.49   & 16   & 0    \\
\midrule
\multicolumn{5}{l}{\emph{Exp. 3: cache size} (Numba, hops $k=2$, batch size=50)} \\ \midrule
cache size  & 1   & 441870.19 & 3010 & 3009 \\
cache size  & 2   & 419436.59 & 2918 & 2916 \\
cache size  & 4   & 384086.42 & 2659 & 2655 \\
cache size  & 8   & 304066.00 & 1967 & 1959 \\
cache size  & 16  & 4037.69   & 16   & 0    \\
\midrule
\multicolumn{5}{l}{\emph{Exp. 4: backend} (cache size $n=16$, hops $k=2$, batch size=50)} \\ \midrule
backend     & Numba     & 4143.16   & 16 & 0 \\
backend     & Scipy     & 3889.86   & 16 & 0 \\
backend     & Torch-CPU & 245311.21 & 16 & 0 \\
backend     & Torch-GPU & 6409.32   & 16 & 0 \\
\bottomrule \bottomrule
\end{tabular}
}
\caption{\small Scalability of LogosKG-Large on the PKG across hops, batch sizes, cache sizes, and backends.}
\label{tab:logoskg_scalability}
\end{table}

\section{Dataset and Setup} 

Our results are organized into two parts: (1) \textsc{LogosKG} system evaluation, focusing on retrieval accuracy, efficiency, and scalability; and (2) an analysis of KG-LLM interaction in high-hop KG regimes, using clinical diagnosis prediction as a representative problem domain. 


We examine three biomedical KGs in our experiments: \textbf{UMLS}~\citep{bodenreider2004unified}, a large-scale biomedical ontology comprising 407K nodes and 3.4M edges across 133 semantic types; the \textbf{PubMed Knowledge Graph (PKG)}~\citep{xu2020building,xu2025pubmed}, a massive citation network connecting 54.4M nodes (including authors, publications, and institutions) via 86.5M edges; and \textbf{PrimeKG}~\citep{chandak2022building}, which integrates 20 high-quality biomedical resources to describe 17,080 diseases with $\approx$4M relationships.

We include two clinical datasets that connect the KGs under study to real-world diagnostic applications. ProbSum \citep{gao-etal-2023-overview} contains de-identified clinical notes with findings and diagnoses, supplying entity mentions that serve as realistic query inputs for retrieval evaluation. DDXPlus \citep{fansi2022ddxplus} offers large-scale symptom-diagnosis pairs and is used in the KG-LLM study to assess how KG structure aligns with LLM-predicted diagnoses. Retrieval experiments use UMLS and PKG with ProbSum, while the KG-LLM case study uses UMLS and PrimeKG with ProbSum and DDXPlus. 

For the second part of the paper where we explore KG-LLMs, we include the following instruction-tuned, widely used LLMs: Qwen-2.5-7B-Instruct \citep{qwen2025qwen25technicalreport}, Llama-3.1-8B-Instruct \citep{dubey2024llama}, GPT 4.1 \citep{openai_gpt4_1}, and GPT-5-mini \citep{openai_gpt5_mini}. We employ GPT-5-mini as the LLM-as-a-Judge and for the supplementary experiments detailed in Appendix~\ref{appendix:refinement}.
All open-sourced LLMs are running on the GPU server. We use GPT-4.1 and GPT-5-mini via HIPAA-compliant Microsoft Azure OpenAI endpoint, thereby complying with the ProbSum Data Use Agreement for MIMIC-III derived content~\cite{johnson2016mimic}.

\section{\textsc{LogosKG} System Evaluation}
\label{sec:logoskg_results}

We evaluate LogosKG and baselines along three dimensions: 
\textbf{Accuracy} measures whether retrieved entities match ground truth across hops; \textbf{Efficiency} is assessed by query time and timeout rates; \textbf{Scalability} examines how performance changes of LogosKG under different settings, evaluating the ability to handle large KGs.

\vspace{.06in}

\noindent \textbf{Baselines}
We compare \textsc{LogosKG} against a diverse set of widely used graph retrieval systems, including database engines (Neo4j), graph analysis toolkits (igraph, NetworkX, graph-tool, SNAP), matrix-based libraries (GraphBLAS and its Python implementation), and GPU-accelerated frameworks (cuGraph, DGL, PyG).
Together, these baselines cover the major design paradigms for KG retrieval, allowing comprehensive evaluation of efficiency, scalability, and retrieval fidelity. 

\noindent \textbf{Accuracy} On a set of queries, we compare the retrieved results from \textsc{LogosKG} family with CPU- and GPU-based baselines across 1–5 hops. This result is measured by Jaccard similarity, with 1.0 denoting exact agreement. All comparisons yield a perfect Jaccard score of 1.0, confirming that retrieval is deterministic given the KG structure. We neglect the detailed results here and present them in Table~\ref{tab:kg-accuracy-family}. This validates \textsc{LogosKG} as a faithful and reliable replacement for existing engines. 


\noindent \textbf{Efficiency}
For each method, we report the average \textbf{query time (QT)} across all test samples and the \textbf{timeout rate (TR)} per hop, defined as the proportion of queries exceeding the hop-specific limit. QT reflects the latency of a method, while TR captures its responsiveness under practical constraints. These metrics provide a clear view of efficiency between \textsc{LogosKG} and the baselines. 

Table~\ref{tab:kg_retrieval_h1to5} reports QT and TR for 1--5 hops. Classical CPU libraries (NetworkX, igraph, SNAP, GraphBLAS) perform well at shallow hops but degrade rapidly with depth. Neo4j remains slow with consistently high TR. GPU-based methods (graph-tool, cuGraph, DGL, PyG) scale roughly linearly: graph-tool, cuGraph, and DGL show higher latency, while PyG starts lower. In contrast, the \textsc{LogosKG} family scales robustly: Numba and Torch-GPU remain under 200 ms with zero TR, and partitioned LogosKG-Large variants maintain moderate TR even at deeper hops. Torch-CPU performs poorly due to the lack of dedicated optimization. Overall, \textsc{LogosKG} achieves efficient and reliable multi-hop retrieval. 

\noindent \textbf{Scalability}
We evaluate LogosKG-Large on the PKG using the degree-aware partitioning strategy. Synthetic queries contain 1--20 randomly selected entities and are executed at depths of 1--5 hops. The cache size $N$ determines how many subgraphs remain in memory, while loads and evictions record swaps between disk and memory. We measure QT under varying hops, batch sizes, cache sizes, and backends. Cache activities are tracked through the number of loads and evictions.

As shown in Table~\ref{tab:logoskg_scalability}, QT increases with hop count as deeper expansions involve more entities, while larger batches reduce latency due to batch optimization. Cache size has the largest impact: small caches trigger frequent swaps, increasing latency, whereas larger caches cut these operations and improve QT. Among backends, Numba, SciPy, and Torch-GPU are fastest, while Torch-CPU remains slower due to limited parallelization.  

\noindent \textbf{Overall performance} These experiments confirm that \textsc{LogosKG} delivers deterministic accuracy, strong efficiency across CPU and GPU settings, and scalable performance on billion-edge graphs, establishing it as a reliable and hardware-optimized solution for large-scale KG retrieval.

\section{Interaction Regimes between High-Hop KGs and LLM for Diagnosis} 
\label{sec:interaction_regimes}
\begin{figure}[t]
    \centering
    \includegraphics[width=0.98\linewidth]{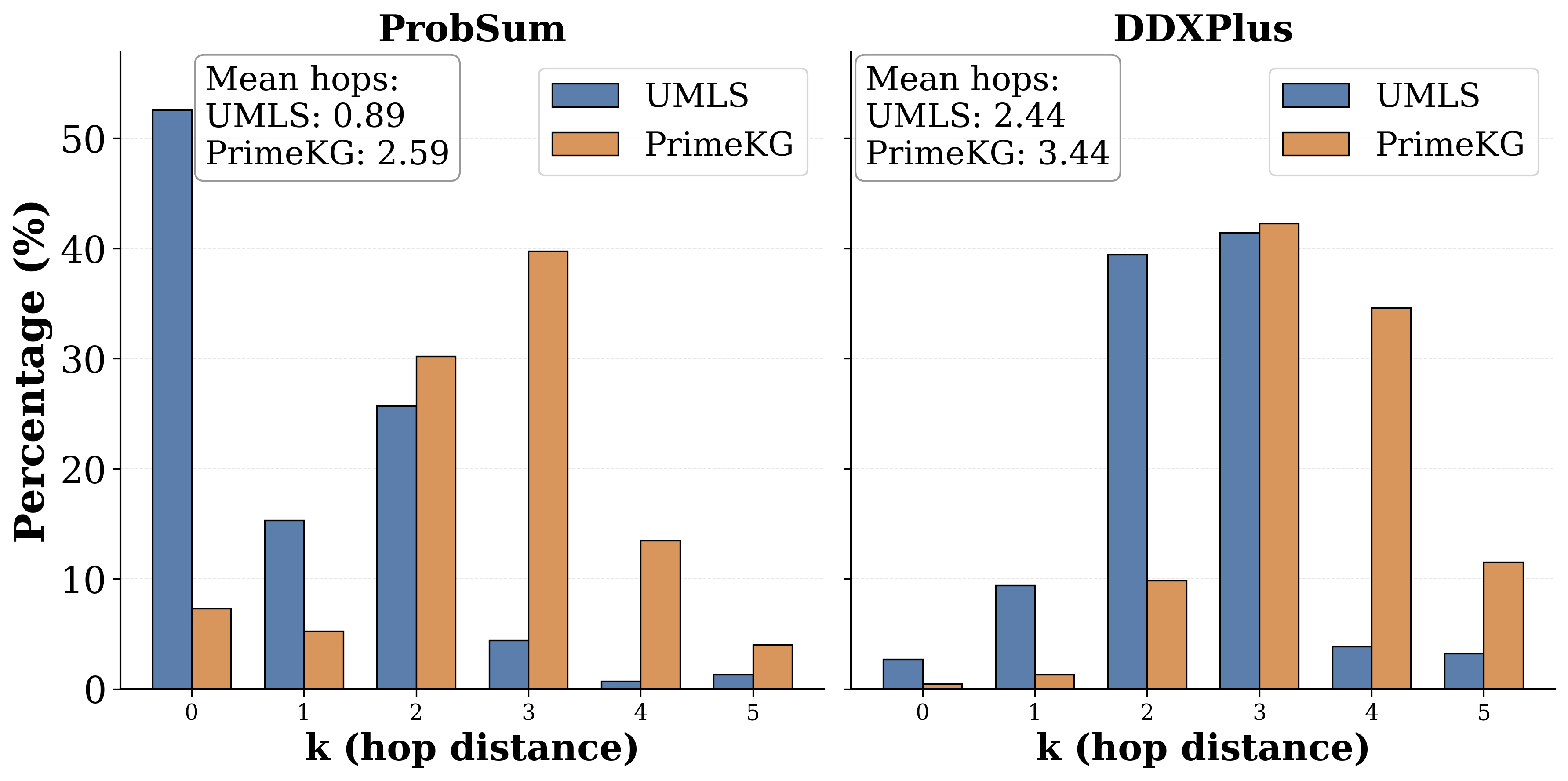}
    \caption{\small Hop-distance distribution of pair entities for the ProbSum and DDXPlus dataset on UMLS and PrimeKG.}
    \label{fig:prob-hop-dist}
\end{figure}

\begin{figure*}[t]
    \centering
    \includegraphics[width=0.98\textwidth]{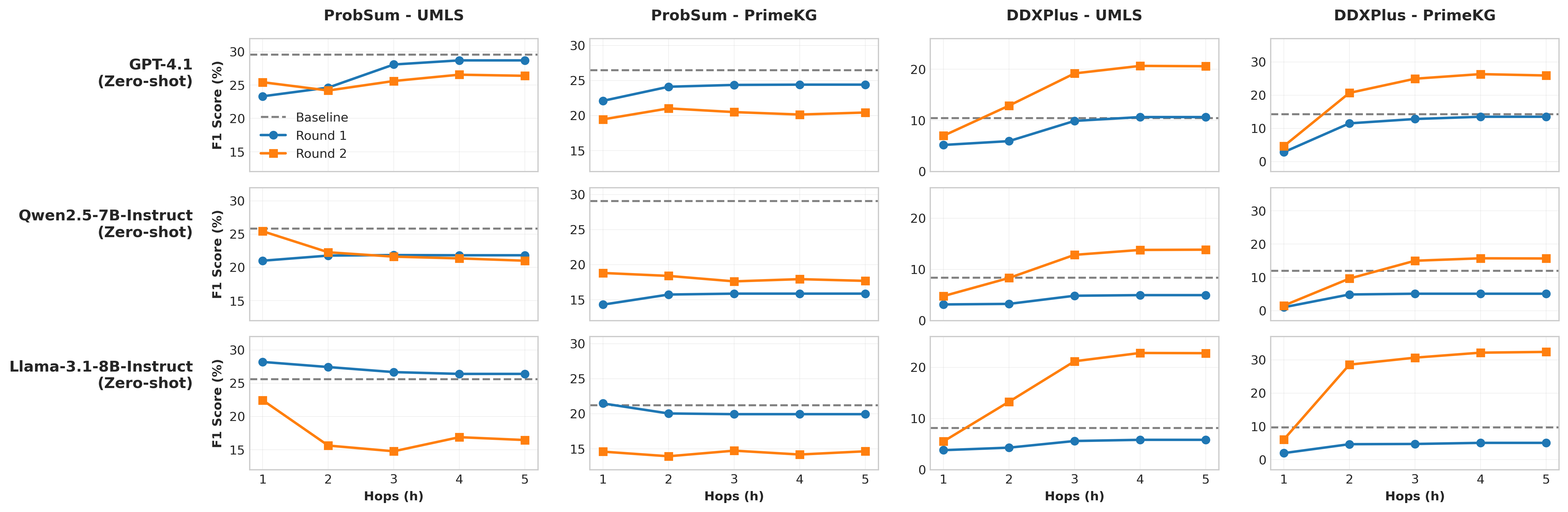}
    \caption{\small Performance comparison of KG filtering (Round 1) and enhancement (Round 2) across varying hop distances ($k=1$ to $k=5$) on ProbSum and DDXPlus datasets using UMLS and PrimeKG. F1-score metrics are shown for \textbf{GPT-4.1, Qwen-2.5-7B-Instruct, and Llama-3.1-8B-Instruct} in the zero-shot setting, with baseline performance included for reference.}
    \label{fig:f1_by_hops_zero_shot}
\end{figure*}

\begin{figure}[t]
    \centering
    \includegraphics[width=0.98\linewidth]{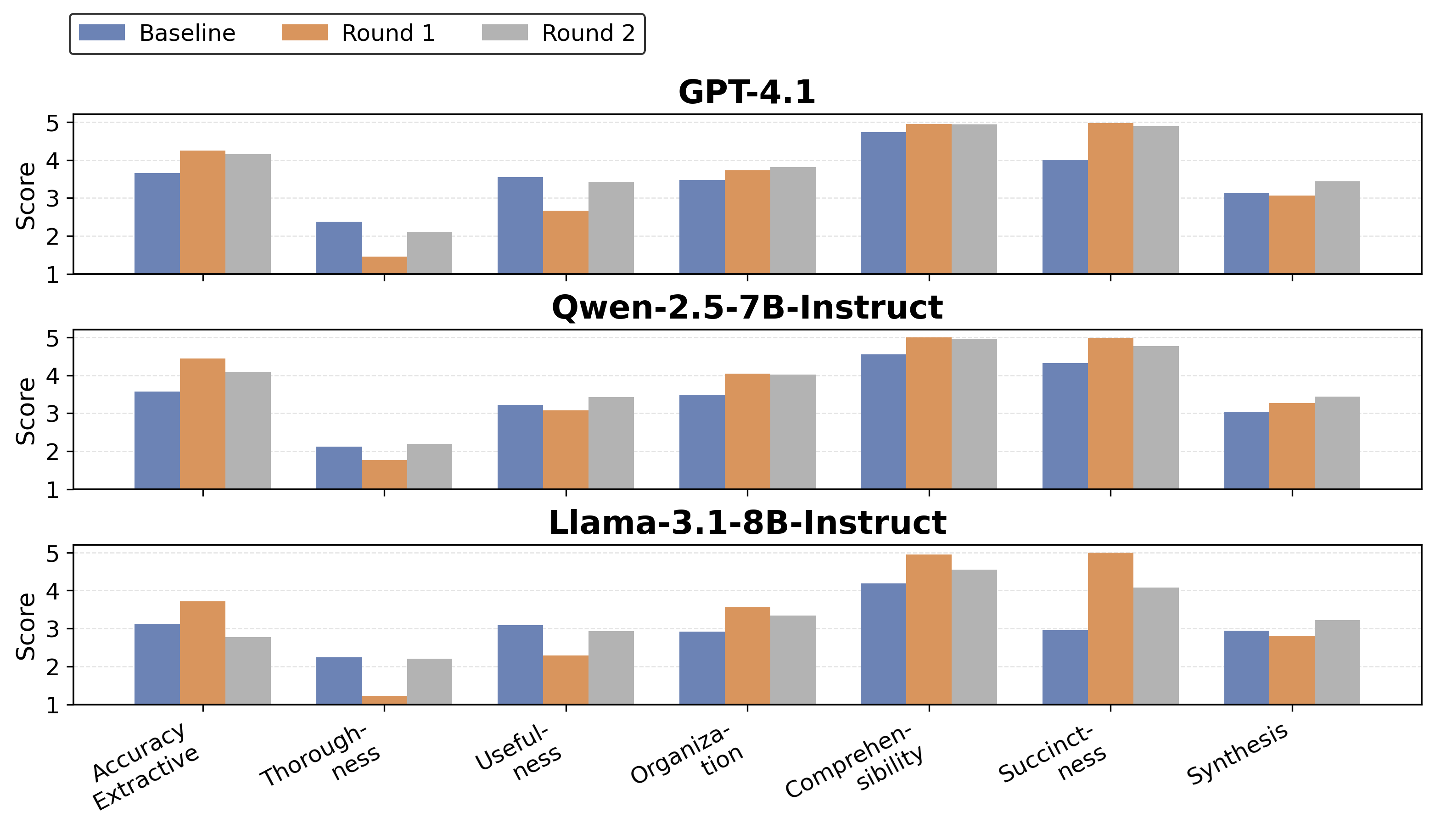}
    \caption{\small PDSQI-9 comparison for three models on DDXPlus dataset with UMLS ($k$=5). Each subplot displays seven evaluation dimensions comparing Baseline, Round 1 (KG-filtered), and Round 2 (KG-enhanced) approaches.}
    \label{fig:pdsqi9_ddxplus}
\end{figure}


Existing approaches that integrate KGs with LLMs often restrict reasoning to shallow neighborhoods due to the exponential growth of the search space. However, many real-world reasoning tasks require traversing distal, multi-hop associations that lie beyond these local regions. In the biomedical KG-LLM setting, for instance, prior work often stops at 1-2 hops (and at most 3 hops) as the higher-hop expansion quickly becomes computationally infeasible~\cite{gao2025leveraging,zuo2025kg4diagnosis,jiang2024graphcare}. Our work deviates from the ``learning-to-rank" paradigm to explore a structural regime of scalable graph traversal, enabled by \textsc{LogosKG}, which makes high-hop reasoning systematically accessible to LLMs. Following prior work~\cite{gao2025leveraging,zuo2025kg4diagnosis}, we use clinical diagnosis as a representative setting to characterize how knowledge graph topology induces distinct interaction regimes between KGs and LLM reasoning.



\textbf{Topological landscape.} Figure~\ref{fig:prob-hop-dist} presents hop distribution across the two KGs on the two datasets we are investigating. On UMLS, gold entities can mostly be found within 1--2 hops, while PrimeKG shows a more long-tail distribution up to 4--5 hops. These structural differences directly govern how KG ontologies interact with LLM predictions.   

\textbf{The precision-recall tradeoffs.}  We adopt a two-round interaction paradigm to understand the interaction between KG topology and systematic precision-recall behaviors of LLM. In Round 1, the KG acts as a structural constraint that eliminates LLM hallucinations. Because LogosKG can retrieve these neighbors with 100\% deterministic fidelity ($Jaccard = 1.0$), the LLM is forced to align its prediction with the medical ontology. In Round 2, \textsc{LogosKG} provides the LLM with a long-tail candidate space that was previously invisible, making the optimal ``depth'' not a fixed parameter. 


As shown in Figure~\ref{fig:f1_by_hops_zero_shot}, for ProbSum, performance changes are relatively moderate across hop distances, with mostly limited gains from KG filtering and enhancement at early hops. In contrast, DDXPlus shows much stronger sensitivity to hop expansion: performance generally improves as the hop distance increases, and Round 2 often provides clear gains beyond Round 1, especially with PrimeKG. Importantly, this pattern holds across all three LLMs, despite differences in absolute performance, indicating that the observed regimes are driven by data and KG structure rather than by model-specific artifacts. These regimes are not tied to a particular retrieval implementation; rather, they were previously obscured because learning-based models and classical graph libraries encounter a computational barrier at high hop distances. By enabling scalable and deterministic high-hop traversal, \textsc{LogosKG} serves as a \textit{structural backbone} that exposes a broader capability space for KG-LLM interaction.

\textbf{Clinical reasoning quality.} We further compare the diagnoses with and without LogosKG ($k$=5) using the clinically validated PDSQI-9 rubric. PDSQI-9 is a nine-criterion framework that evaluates diagnostic documentation along dimensions such as accuracy, comprehensibility, organization, and succinctness, with a HIPAA-compliant Azure GPT-5-mini model serving as the LLM-as-a-Judge~\cite{croxford2025development,croxford2025evaluating}. As shown in Figure~\ref{fig:pdsqi9_ddxplus}, incorporating \textsc{LogosKG} leads to consistent improvements across several dimensions, particularly accuracy extractive, organization, comprehensibility, succinctness, and synthesis. These gains primarily reflect changes in the structure and clarity of diagnostic reasoning rather than raw correctness alone. By constraining predictions to KG-supported evidence in Round 1, \textsc{LogosKG} suppresses unsupported or redundant diagnoses, which directly benefits accuracy extractive, organization, comprehensibility, and succinctness. Round 2 further improves synthesis by reintroducing clinically relevant conditions that the LLM may omit under unconstrained generation. Improvements are observed consistently for all models, and are especially pronounced on the DDXPlus dataset, where deeper relational contexts are required. Full results are provided in Appendix \S\ref{appendix:pdsqi9_full}.

\begin{figure*}[t]
    \centering
    \includegraphics[width=0.98\textwidth]{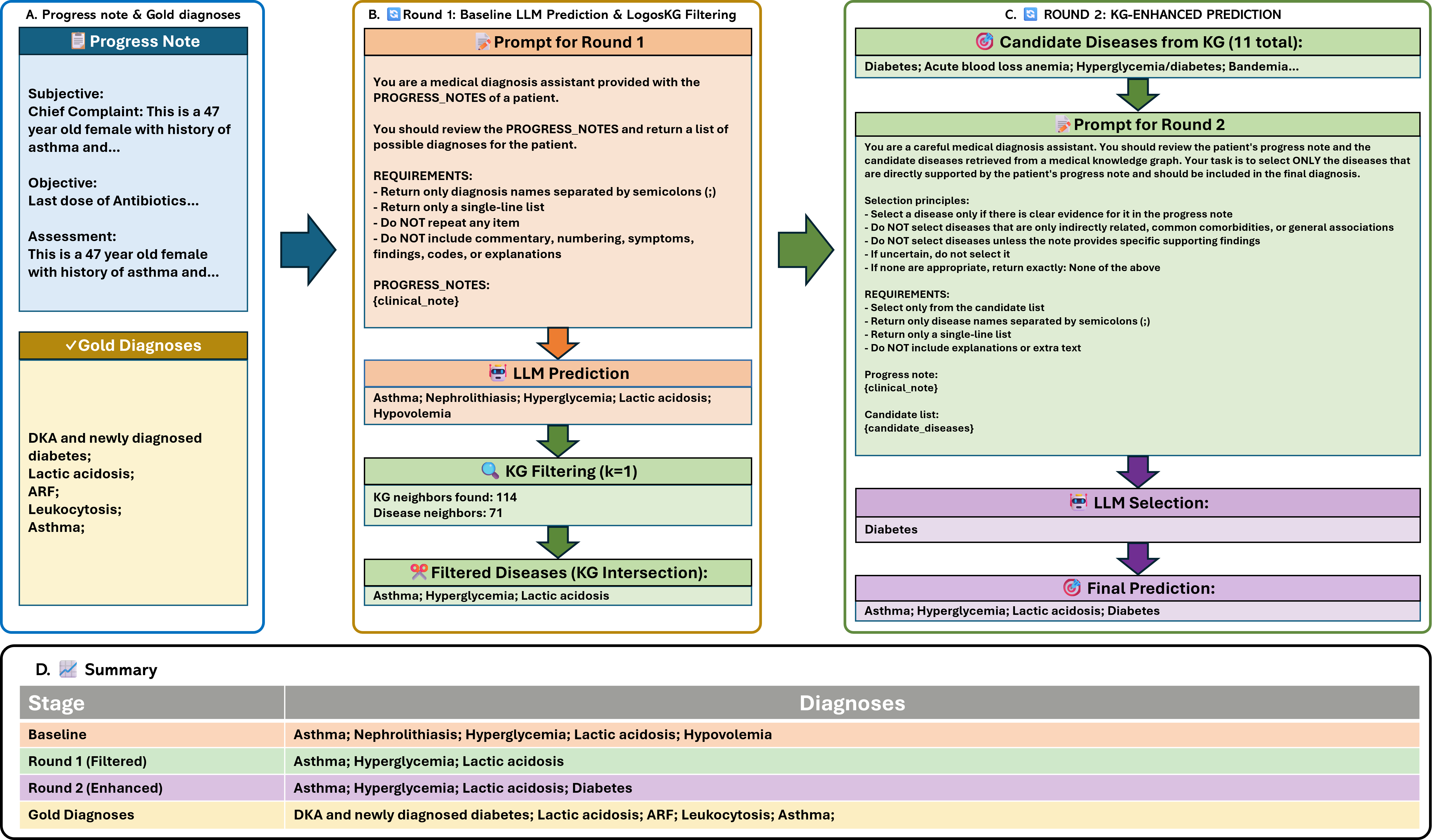}
    \caption{Illustrative example of the \textsc{LogosKG} two-round prediction pipeline on a clinical case from ProbSum. \textbf{(A)} Clinical input consists of a patient progress note with the gold standard diagnosis. \textbf{(B)} Round 1: The baseline LLM generates differential diagnoses, which are then filtered by validating against KG neighbors retrieved within $k=1$ hop. \textbf{(C)} Round 2: Additional candidate diseases from the KG are presented to the LLM for selection, and the final prediction combines the filtered Round 1 diagnoses with the disease selected in Round 2. \textbf{(D)} Summary table comparing baseline, Round 1 (KG-filtered), Round 2 (KG-enhanced), and gold standard diagnoses.}
    \label{fig:logoskg_example}
\end{figure*}

\textbf{Additional analysis.} We provide results from fine-tuned LLMs and similarity-based refinement built on \textsc{LogosKG} in the Appendix (Figure~\ref{fig:prf1_by_hops_sft} and Table~\ref{tab:retrieval_refinement}) for interested readers. While using LogosKG for high-hop retrieval followed by a fine-tuned selector, or even a multi-agent workflow, is an extension, our focus here is to analyze how \textsc{LogosKG} enables and characterizes the underlying structural interaction regimes because of its efficiency and scalability.

\subsection{Case Study of LLM Diagnoses with \textsc{LogosKG} Filtering and Enhancement}
Figure~\ref{fig:logoskg_example} shows how \textsc{LogosKG} works in practice on a real clinical case. The baseline LLM generates five diagnoses from the patient's progress note, namely \textsc{Asthma, Nephrolithiasis, Hyperglycemia, Lactic acidosis}, and \textsc{Hypovolemia}. When we filter these with the KG in Round 1, \textsc{Asthma, Hyperglycemia} and \textsc{Lactic acidosis} remain, while the others fail to connect to clinical entities extracted from the note within $k=1$ hop. This strict filtering removes unsupported predictions but also removes potentially relevant diagnoses.

Round 2 addresses this limitation by finding eleven candidate diseases from the KG based on the patient's clinical findings. We then ask the LLM to select the most relevant candidates from this larger set. The model identifies one important disease: \textsc{Diabetes}. Notably, this matches well with the gold standard, which includes \textsc{DKA and newly diagnosed diabetes}, even though it does not appear in the baseline prediction. 

The final output combines the Round 1 filtered baseline diagnoses with the disease selected in Round 2, achieving much better coverage than either the baseline alone or the Round 1 result. This example also illustrates the motivation of proposing \textsc{LogosKG}: KGs can expand the search space beyond what LLMs initially consider, while LLM reasoning helps select which candidates are actually relevant to the patient.



\section{Conclusion}
We presented \textsc{LogosKG}, which is a hardware-aligned framework for scalable and interpretable multi-hop retrieval on large KGs. \textsc{LogosKG} achieves efficient and deterministic retrieval across billion-scale graphs. Our analysis shows that KG topology strongly shapes how structured knowledge interacts with LLM reasoning, positioning \textsc{LogosKG} as both a high-performance retrieval system and a foundation for future research on KG-LLM integration. 

\section*{Acknowledgments}
This work is supported by U.S. National Library of Medicine, National Institute of Health, under award number R00LM014308.



\section*{Limitations}
While LogosKG focuses on efficient multi-hop retrieval, the current implementation can be further developed to support other types of graph analysis based on similar graph reasoning operations. Future extensions may explore broader algorithmic capabilities while maintaining efficiency and scalability.  
In addition, retrieval from large KGs often produces a substantial number of candidate entities at higher hop distances. Further work is needed to refine these outputs, for example, by integrating more effective ranking or filtering strategies to highlight the most relevant results.

Finally, \textsc{LogosKG} serves as a helpful graph retrieval tool, but it does not guarantee consistent performance improvements in every case, as success depends on both the completeness of the KG and the reasoning of the LLM. In Round 1, we are limited by the graph's coverage; if the KG lacks evidence for a specific symptom-diagnosis pair, retrieval offers little benefit. Similarly, Round 2 relies on the LLM accurately reselecting information. If the model makes incorrect choices or hallucinates during this step, the potential for improvement is limited by the LLM's own reasoning capabilities. 



\section*{Ethical Statement}
This study does not involve any human subjects. All datasets and models used, including UMLS, PKG, PrimeKG, ProbSum, and DDXPlus, are publicly available and used strictly for research purposes under their respective data use agreements. All experiments were conducted on HIPAA-compliant servers, adhering to institutional and data governance policies.

\bibliography{custom}

\appendix


\appendix
\section{Appendix}
\label{sec:appendix}


\subsection{Jaccard similarity for LogosKG retrieval fidelity}

To evaluate retrieval accuracy, we use the Jaccard similarity:
{\small 
\[
\mathrm{Jaccard}(A,B) = \frac{|R_A \cap R_B|}{|R_A \cup R_B|},
\]
}
where $R_A$ and $R_B$ are the retrieved entity sets from the method $A$ and method $B$ on the same query. Higher values indicate stronger agreement (1.00 = identical results).

\begin{table}[h]
\centering
\resizebox{\linewidth}{!}{
\begin{tabular}{lccccc}
\toprule \toprule
\textbf{LogosKG family} & \textbf{Hop 1} & \textbf{Hop 2} & \textbf{Hop 3} & \textbf{Hop 4} & \textbf{Hop 5} \\
\midrule
Neo4j        & 1.00 & 1.00 & 1.00 & 1.00 & 1.00 \\
GraphBLAS    & 1.00 & 1.00 & 1.00 & 1.00 & 1.00 \\
igraph       & 1.00 & 1.00 & 1.00 & 1.00 & 1.00 \\
NetworkX     & 1.00 & 1.00 & 1.00 & 1.00 & 1.00 \\
graph-tool   & 1.00 & 1.00 & 1.00 & 1.00 & 1.00 \\
SNAP         & 1.00 & 1.00 & 1.00 & 1.00 & 1.00 \\
\midrule
cuGraph      & 1.00 & 1.00 & 1.00 & 1.00 & 1.00 \\
DGL          & 1.00 & 1.00 & 1.00 & 1.00 & 1.00 \\
PyG          & 1.00 & 1.00 & 1.00 & 1.00 & 1.00 \\
\bottomrule \bottomrule
\end{tabular}
}
\caption{Jaccard similarity of the \textbf{LogosKG family} vs.\ CPU and GPU baselines across hops 1--5.}
\label{tab:kg-accuracy-family}
\end{table}

Table~\ref{tab:kg-accuracy-family} shows that all retrieved results are perfectly overlapped, proving the fidelity of \textsc{LogosKG}. 

\begin{figure*}[t]
    \centering
    \includegraphics[width=0.94\textwidth]{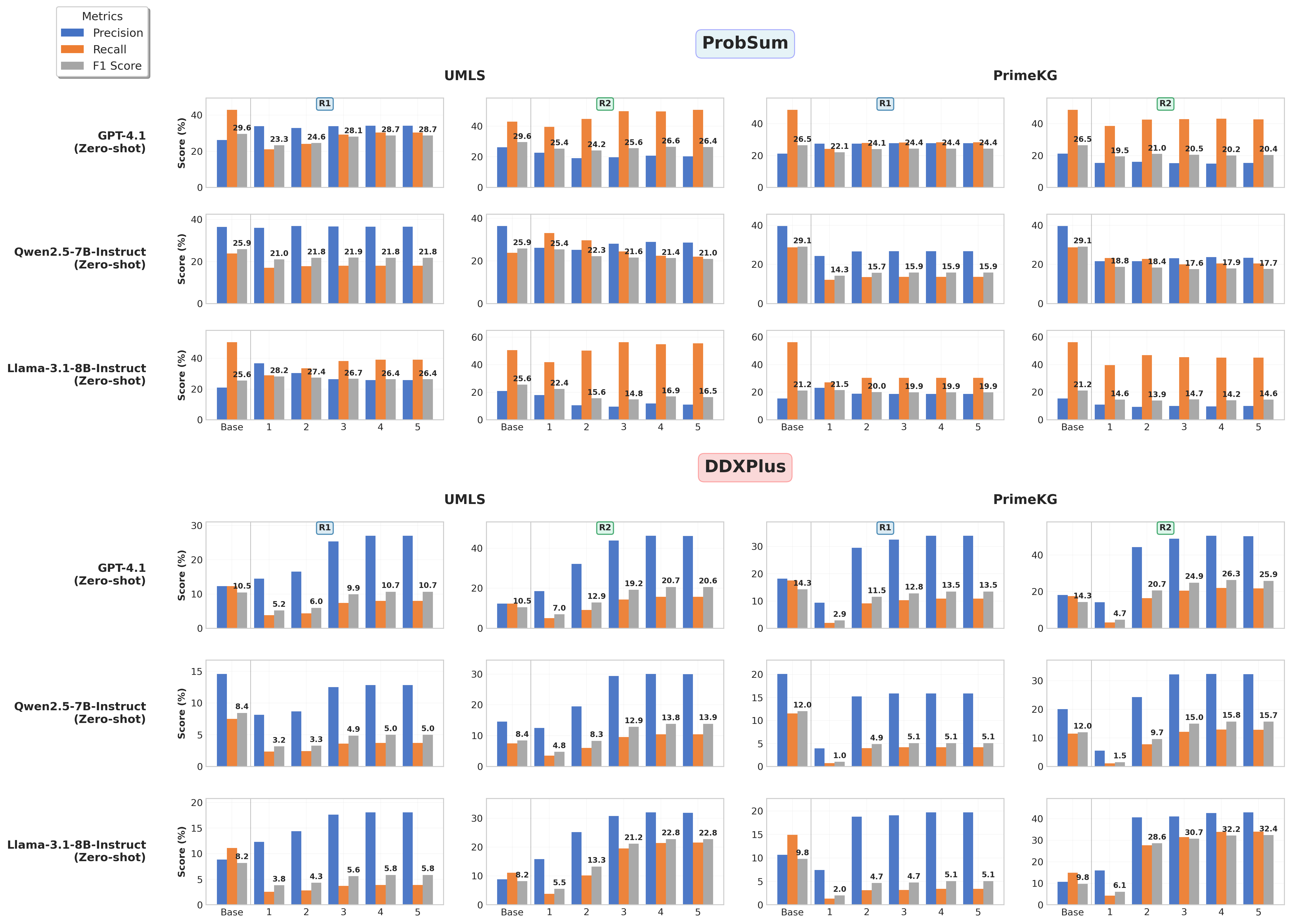}
    \caption{Performance comparison of KG filtering (Round 1) and enhancement (Round 2) across varying hop distances ($k=1$ to $k=5$) on ProbSum and DDXPlus datasets using UMLS and PrimeKG. Precision, recall, and F1 score metrics are shown for \textbf{GPT-4.1, Qwen-2.5-7B-Instruct, and Llama-3.1-8B-Instruct} in the Zero-shot setting, with baseline performance included for reference.}
    \label{fig:prf1_by_hops_zero_shot}
\end{figure*}

\subsection{Entity-level Evaluation Metric}
\label{sc:entity_evaluation}

We compute the baseline precision and recall based on the overlap between predicted and gold-standard entities, as defined below:

{\small
\begin{equation}
\label{eq:precision_recall}
\begin{aligned}
\mathrm{Precision} &= \frac{|\,pred \cap gold\,|}{|\,pred\,|},\\
\mathrm{Recall} &= \frac{|\,pred \cap gold\,|}{|\,gold\,|}.
\end{aligned}
\end{equation}
}



We compute precision and recall to assess the effect of KG processing. Round 1 (filtered) and Round 2 (enhanced) use the same computation, which can be formulated as follows. Let $pred_f$ denote the subset of predictions after KG processing (filtered in Round 1, enhanced in Round 2):

{\small
\begin{equation}
\label{eq:kg_precision_recall}
\begin{aligned}
\mathrm{Precision}_f &= \frac{|\,pred_f \cap gold\,|}{|\,pred_f\,|},\\
\mathrm{Recall}_f &= \frac{|\,pred_f \cap gold\,|}{|\,gold\,|}.
\end{aligned}
\end{equation}
}

\subsection{Supplementary Sensitivity Analysis}

We study the impact of \textsc{LogosKG} on LLM performance across varying retrieval depths ($k=1$--$5$) in Rounds 1 and 2. Figure \ref{fig:prf1_by_hops_zero_shot} details precision, recall, and F1 scores, supplementing the main results in Figure \ref{fig:f1_by_hops_zero_shot}. 

In Round 1, increasing the retrieval depth allows \textsc{LogosKG} to uncover more supporting evidence from KGs. We observe this as an increase in recall; however, due to the inherent incompleteness of the KG, recall remains below the baseline. Conversely, precision consistently exceeds the baseline, as \textsc{LogosKG} effectively filters out hallucinated diagnoses using retrieval evidence. To recover important diagnoses missed by the LLM, Round 2 prompts the model to select additional evidence from the \textsc{LogosKG} retrieval evidence. Here, we observe that recall surpasses the baseline, though this comes with a slight decrease in precision as the inclusion of more evidence inevitably introduces some noise. Nevertheless, the overall improvement in F1 scores demonstrates that \textsc{LogosKG} successfully enhances diagnostic performance. These results underscore the critical role that KG structure plays in determining how much LLMs can benefit from integration.

\subsection{Diagnosis Quality Evaluation using PDSQI-9 Criteria}
\label{appendix:pdsqi9_full}

\begin{table*}[!h]
\centering
\resizebox{0.94\textwidth}{!}{%
\begin{tabular}{cccccccccccccccccccc}
\toprule
\multirow{3}{*}[-5pt]{\textbf{Dataset}} & \multirow{3}{*}[-5pt]{\textbf{Metric}} & \multicolumn{9}{c}{\textbf{UMLS}} & \multicolumn{9}{c}{\textbf{PrimeKG}} \\
\cmidrule(lr){3-11}  \cmidrule(lr){12-20}
 &  & \multicolumn{3}{c}{\textbf{GPT-4.1}} & \multicolumn{3}{c}{\textbf{Qwen-2.5-7B-Instruct}} & \multicolumn{3}{c}{\textbf{Llama-3.1-8B-Instruct}} & \multicolumn{3}{c}{\textbf{GPT-4.1}} & \multicolumn{3}{c}{\textbf{Qwen-2.5-7B-Instruct}} & \multicolumn{3}{c}{\textbf{Llama-3.1-8B-Instruct}} \\
 \cmidrule(lr){3-5} \cmidrule(lr){6-8} \cmidrule(lr){9-11}  \cmidrule(lr){12-14} \cmidrule(lr){15-17} \cmidrule(lr){18-20}
 &  & B & R1 & R2 & B & R1 & R2 & B & R1 & R2 & B & R1 & R2 & B & R1 & R2 & B & R1 & R2 \\
\midrule
\multirow{8}{*}{\textbf{ProbSum}} & Accuracy extractive & 4.29 & 3.86 & 1.84 & 4.07 & 4.11$\uparrow$ & 3.29 & 2.76 & 3.36$\uparrow$ & 1.32 & 4.22 & 3.90 & 2.10 & 4.00 & 4.08$\uparrow$ & 3.17 & 2.66 & 3.17$\uparrow$ & 1.39 \\
 & Thoroughness & 3.00 & 1.25 & 2.05 & 1.61 & 1.22 & 1.40 & 3.04 & 1.59 & 1.51 & 2.90 & 1.24 & 2.05 & 1.57 & 1.24 & 1.35 & 3.09 & 1.30 & 1.54 \\
 & Usefulness & 4.01 & 2.18 & 2.02 & 2.80 & 2.21 & 2.20 & 2.98 & 2.53 & 1.30 & 3.96 & 2.00 & 2.29 & 2.92 & 2.12 & 2.10 & 2.92 & 2.18 & 1.49 \\
 & Organization & 3.57 & 3.18 & 2.02 & 3.53 & 3.28 & 2.85 & 2.70 & 3.16$\uparrow$ & 1.53 & 3.62 & 3.17 & 2.23 & 3.40 & 3.31 & 2.78 & 2.75 & 3.14$\uparrow$ & 1.63 \\
 & Comprehensibility & 4.91 & 4.91 & 2.68 & 4.74 & 4.92$\uparrow$ & 4.18 & 3.90 & 4.52$\uparrow$ & 2.34 & 4.90 & 4.92$\uparrow$ & 2.77 & 4.65 & 4.88$\uparrow$ & 3.86 & 3.85 & 4.54$\uparrow$ & 2.44 \\
 & Succinctness & 4.59 & 5.00$\uparrow$ & 1.87 & 4.83 & 5.00$\uparrow$ & 3.87 & 2.44 & 4.66$\uparrow$ & 1.58 & 4.51 & 5.00$\uparrow$ & 2.43 & 4.71 & 5.00$\uparrow$ & 3.77 & 2.34 & 4.77$\uparrow$ & 1.78 \\
 & Synthesis & 3.33 & 2.86 & 2.15 & 3.12 & 2.89 & 2.49 & 2.84 & 2.99$\uparrow$ & 1.46 & 3.36 & 2.74 & 2.30 & 3.15 & 2.94 & 2.54 & 2.84 & 2.89$\uparrow$ & 1.54 \\
 & \textit{Average} & 3.96 & 3.32 & 2.09 & 3.53 & 3.38 & 2.90 & 2.95 & 3.26$\uparrow$ & 1.58 & 3.92 & 3.28 & 2.31 & 3.49 & 3.37 & 2.80 & 2.92 & 3.14$\uparrow$ & 1.69 \\
\midrule
\multirow{8}{*}{\textbf{DDXPlus}} & Accuracy extractive & 3.66 & 4.24$\uparrow$ & 4.16$\uparrow$ & 3.58 & 4.45$\uparrow$ & 4.08$\uparrow$ & 3.13 & 3.72$\uparrow$ & 2.77 & 3.65 & 4.38$\uparrow$ & 4.11$\uparrow$ & 3.72 & 4.61$\uparrow$ & 3.66 & 3.00 & 3.68$\uparrow$ & 2.76 \\
 & Thoroughness & 2.38 & 1.46 & 2.11 & 2.12 & 1.77 & 2.20$\uparrow$ & 2.24 & 1.22 & 2.20 & 2.36 & 1.70 & 2.17 & 2.14 & 1.81 & 2.12 & 2.30 & 1.44 & 2.12 \\
 & Usefulness & 3.54 & 2.67 & 3.42 & 3.22 & 3.08 & 3.42$\uparrow$ & 3.08 & 2.29 & 2.93 & 3.46 & 3.01 & 3.56$\uparrow$ & 3.20 & 3.32$\uparrow$ & 3.19 & 2.94 & 2.44 & 2.91 \\
 & Organization & 3.48 & 3.73$\uparrow$ & 3.81$\uparrow$ & 3.49 & 4.04$\uparrow$ & 4.02$\uparrow$ & 2.92 & 3.56$\uparrow$ & 3.34$\uparrow$ & 3.57 & 4.01$\uparrow$ & 4.07$\uparrow$ & 3.59 & 4.06$\uparrow$ & 3.95$\uparrow$ & 2.94 & 3.67$\uparrow$ & 3.38$\uparrow$ \\
 & Comprehensibility & 4.73 & 4.96$\uparrow$ & 4.93$\uparrow$ & 4.55 & 5.00$\uparrow$ & 4.96$\uparrow$ & 4.19 & 4.95$\uparrow$ & 4.55$\uparrow$ & 4.72 & 4.99$\uparrow$ & 4.85$\uparrow$ & 4.61 & 5.00$\uparrow$ & 4.91$\uparrow$ & 4.11 & 4.88$\uparrow$ & 4.52$\uparrow$ \\
 & Succinctness & 4.01 & 4.98$\uparrow$ & 4.89$\uparrow$ & 4.33 & 4.99$\uparrow$ & 4.77$\uparrow$ & 2.96 & 5.00$\uparrow$ & 4.08$\uparrow$ & 3.89 & 4.99$\uparrow$ & 4.82$\uparrow$ & 4.28 & 5.00$\uparrow$ & 4.53$\uparrow$ & 2.72 & 4.98$\uparrow$ & 4.18$\uparrow$ \\
 & Synthesis & 3.13 & 3.07 & 3.44$\uparrow$ & 3.04 & 3.26$\uparrow$ & 3.44$\uparrow$ & 2.95 & 2.80 & 3.22$\uparrow$ & 3.17 & 3.18$\uparrow$ & 3.45$\uparrow$ & 3.12 & 3.28$\uparrow$ & 3.26$\uparrow$ & 2.95 & 3.30$\uparrow$ & 3.22$\uparrow$ \\
 & \textit{Average} & 3.56 & 3.59$\uparrow$ & 3.82$\uparrow$ & 3.47 & 3.80$\uparrow$ & 3.84$\uparrow$ & 3.07 & 3.36$\uparrow$ & 3.30$\uparrow$ & 3.54 & 3.75$\uparrow$ & 3.86$\uparrow$ & 3.52 & 3.87$\uparrow$ & 3.66$\uparrow$ & 2.99 & 3.48$\uparrow$ & 3.30$\uparrow$ \\
\bottomrule
\end{tabular}
}
\caption{PDSQI-9 evaluation comparing Baseline (B), LogosKG Round 1 (R1), and Round 2 (R2) at $k=5$. Results for base models across UMLS and PrimeKG. $\uparrow$ indicates improvement over baseline.}
\label{tab:pdsqi9_comparison}
\end{table*}

\begin{figure*}[!h]
    \centering
    \includegraphics[width=0.94\textwidth]{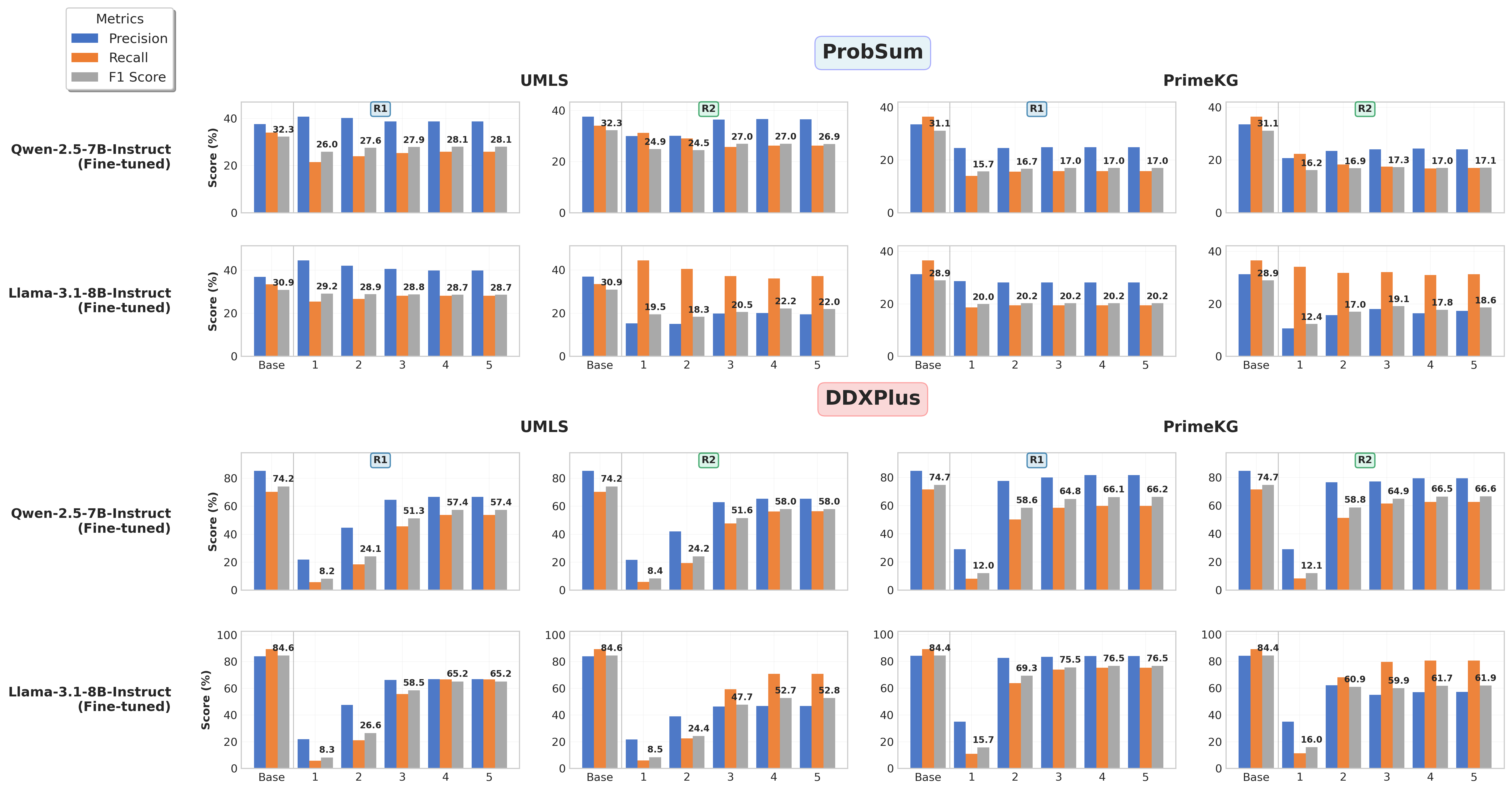}
    \caption{Performance comparison of KG filtering (Round 1) and enhancement (Round 2) across varying hop distances ($k=1$ to $k=5$) on ProbSum and DDXPlus datasets using UMLS and PrimeKG. Precision, recall, and F1 score metrics are shown for \textbf{Qwen-2.5-7B-Instruct (Fine-tuned) and Llama-3.1-8B-Instruct (Fine-tuned)}, with baseline performance included for reference.}
    \label{fig:prf1_by_hops_sft}
\end{figure*}

Table \ref{tab:pdsqi9_comparison} shows that \textsc{LogosKG} generally improves diagnosis quality across models and datasets. For DDXPlus, the gains are particularly evident in accuracy extractive, organization, comprehensibility, succinctness, and synthesis, suggesting that high-hop retrieval yields more concise and comprehensive outputs. Specifically, gains in succinctness are most apparent in Round 1, showing the effectiveness of \textsc{LogosKG} in removing hallucinated diagnoses. In contrast, the gains in comprehensibility and synthesis during Round 2 confirm the system's ability to complement incomplete diagnoses by adding relevant evidence from KGs. Improvements are limited on ProbSum and mainly focused on Round 1, where most concepts are already covered by the input notes (as shown in Figure \ref{fig:prob-hop-dist}), but are much stronger on DDXPlus, which requires retrieving evidence at higher hops. Improvements across multiple metrics confirm that \textsc{LogosKG} effectively enhances LLM diagnostic quality.

\subsection{Analysis: Fine-Tuned Models with KG Enhancement}
\label{appendix:llm_round2_sft}
Figure~\ref{fig:prf1_by_hops_sft} shows how task-specific fine-tuned models perform with \textsc{LogosKG} enhancement across different hop distances. Unlike zero-shot models that gain substantially from KG guidance, fine-tuned models show minimal improvements or even worse performance. Round 1 KG filtering reduces recall while providing only small precision gains, making the trade-off not worthwhile. Round 2 enhancement helps partially but cannot restore baseline performance across both datasets and KGs. This differs sharply from zero-shot scenarios, where the two-round approach successfully balances precision and recall.

\textbf{Why KG Enhancement Cannot Beat Task-Specific Fine-Tuning?} 
This performance gap shows three key problems. First, \textit{knowledge mismatch}: fine-tuned models learn disease connections directly from labeled clinical data, picking up patterns and relationships that may not exist in general biomedical KGs. The KG shows how concepts relate in theory rather than how diseases appear in actual clinical practice. Second, \textit{filtering removes good predictions}: Round 1's strict KG checking throws out predictions that the model correctly learned during training, simply because they don't connect strongly to extracted entities in the KG. Third, \textit{searching space complexity}: While Round 2 finds hundreds of candidate diseases, the model must select from this large set without training.

\textbf{When Does \textsc{LogosKG} Help?}
Our results show that \textsc{LogosKG} works best in zero-shot and few-shot settings where models lack task-specific training. While fine-tuning performs well on target domains, it narrows model knowledge and reduces generalization. \textsc{LogosKG} takes a different approach: it maintains broad capabilities while improving domain-specific performance through external knowledge. This makes it particularly valuable for applications covering diverse medical specialties or rare diseases with limited training data, where preserving model flexibility is as important as achieving strong performance. When task-specific data is scarce or cross-domain generalization is prioritized, KG enhancement offers a practical alternative to fine-tuning.

\begin{table}[!t]
\centering
\resizebox{\columnwidth}{!}{
\begin{tabular}{c l l ccc ccc}
\toprule
\multirow{2}{*}[-2pt]{$k$} & \multirow{2}{*}[-2pt]{Strategy} & \multirow{2}{*}[-2pt]{Setting} & \multicolumn{3}{c}{UMLS} & \multicolumn{3}{c}{PrimeKG} \\
\cmidrule(lr){4-6} \cmidrule(lr){7-9}
 &  &  & P & R & F1 & P & R & F1 \\
\midrule
 & Baseline &  & 17.19 & 46.89 & 23.55 & 13.34 & 47.51 & 19.38 \\
\midrule
\multirow{8}{*}{1} & \multirow{3}{*}{Similarity} & $\tau = 0.7$ & 37.36 & 27.60 & 28.64 & 27.15 & 12.16 & 14.46 \\
 &  & $\tau = 0.8$ & 37.77 & 27.34 & 28.62 & 27.32 & 11.43 & 13.81 \\
 &  & $\tau = 0.9$ & 37.91 & 27.18 & 28.67 & 28.01 & 11.57 & 14.06 \\ \cmidrule{3-9}
 & \multirow{5}{*}{Top-N} & $N = 10$ & 36.58 & 27.37 & 28.18 & 26.94 & 12.06 & 14.22 \\
 &  & $N = 15$ & 36.24 & 28.12 & 28.63 & 26.25 & 12.67 & 14.61 \\
 &  & $N = 20$ & 35.83 & 28.30 & 28.53 & 25.84 & 13.36 & 15.11 \\
 &  & $N = 25$ & 35.62 & 28.36 & 28.47 & 27.25 & 14.03 & 15.81 \\
 &  & $N = 30$ & 35.55 & 28.36 & 28.46 & 26.42 & 14.11 & 15.89 \\
\midrule
\multirow{8}{*}{2} & \multirow{3}{*}{Similarity} & $\tau = 0.7$ & 35.28 & 27.83 & 27.94 & 25.71 & 13.87 & 15.41 \\
 &  & $\tau = 0.8$ & 37.22 & 27.57 & 28.44 & 26.60 & 12.18 & 14.19 \\
 &  & $\tau = 0.9$ & 38.25 & 27.57 & 28.95 & 28.26 & 12.39 & 14.80 \\ \cmidrule{3-9}
 & \multirow{5}{*}{Top-N} & $N = 10$ & 35.65 & 26.58 & 27.28 & 26.92 & 12.12 & 14.30 \\
 &  & $N = 15$ & 34.84 & 28.54 & 28.27 & 25.43 & 13.64 & 15.15 \\
 &  & $N = 20$ & 33.72 & 29.37 & 28.45 & 25.80 & 14.55 & 15.90 \\
 &  & $N = 25$ & 32.65 & 29.62 & 28.14 & 24.19 & 14.67 & 15.70 \\
 &  & $N = 30$ & 32.14 & 30.26 & 28.40 & 23.57 & 14.64 & 15.46 \\
\midrule
\multirow{8}{*}{3} & \multirow{3}{*}{Similarity} & $\tau = 0.7$ & 34.49 & 28.02 & 27.79 & 25.06 & 13.79 & 15.15 \\
 &  & $\tau = 0.8$ & 37.47 & 28.04 & 28.77 & 27.96 & 12.92 & 15.09 \\
 &  & $\tau = 0.9$ & 38.17 & 27.52 & 28.89 & 28.27 & 12.58 & 14.92 \\ \cmidrule{3-9}
 & \multirow{5}{*}{Top-N} & $N = 10$ & 36.73 & 25.42 & 26.83 & 26.35 & 11.07 & 13.57 \\
 &  & $N = 15$ & 33.30 & 27.73 & 27.24 & 27.98 & 13.81 & 15.73 \\
 &  & $N = 20$ & 33.02 & 28.44 & 27.66 & 25.54 & 14.02 & 15.44 \\
 &  & $N = 25$ & 32.38 & 29.48 & 28.10 & 25.29 & 14.47 & 15.71 \\
 &  & $N = 30$ & 31.75 & 29.77 & 27.94 & 24.20 & 14.79 & 15.81 \\
\midrule
\multirow{8}{*}{4} & \multirow{3}{*}{Similarity} & $\tau = 0.7$ & 34.50 & 28.40 & 28.04 & 25.55 & 13.81 & 15.28 \\
 &  & $\tau = 0.8$ & 37.23 & 27.55 & 28.37 & 27.42 & 12.46 & 14.58 \\
 &  & $\tau = 0.9$ & 37.89 & 27.00 & 28.44 & 28.72 & 12.65 & 15.05 \\ \cmidrule{3-9}
 & \multirow{5}{*}{Top-N} & $N = 10$ & 36.91 & 25.83 & 27.12 & 24.92 & 10.59 & 13.03 \\
 &  & $N = 15$ & 33.83 & 27.56 & 27.38 & 26.61 & 13.34 & 15.11 \\
 &  & $N = 20$ & 33.08 & 28.73 & 27.94 & 27.31 & 14.35 & 16.00 \\
 &  & $N = 25$ & 33.15 & 29.60 & 28.35 & 25.79 & 14.35 & 15.67 \\
 &  & $N = 30$ & 31.64 & 29.67 & 27.84 & 25.00 & 14.82 & 15.96 \\
\midrule
\multirow{8}{*}{5} & \multirow{3}{*}{Similarity} & $\tau = 0.7$ & 34.21 & 27.98 & 27.70 & 25.88 & 13.99 & 15.49 \\
 &  & $\tau = 0.8$ & 37.23 & 27.55 & 28.37 & 27.89 & 12.78 & 14.92 \\
 &  & $\tau = 0.9$ & 38.21 & 27.46 & 28.83 & 26.90 & 12.04 & 14.18 \\ \cmidrule{3-9}
 & \multirow{5}{*}{Top-N} & $N = 10$ & 35.85 & 25.29 & 26.52 & 26.08 & 11.17 & 13.65 \\
 &  & $N = 15$ & 33.84 & 27.80 & 27.42 & 26.73 & 13.20 & 14.90 \\
 &  & $N = 20$ & 33.46 & 29.15 & 28.25 & 26.99 & 14.52 & 16.15 \\
 &  & $N = 25$ & 33.14 & 29.64 & 28.37 & 26.15 & 14.53 & 15.89 \\
 &  & $N = 30$ & 31.87 & 29.67 & 27.88 & 25.19 & 14.73 & 15.89 \\
\bottomrule
\end{tabular}
}
\caption{Retrieval refinement performance on UMLS and PrimeKG. Results are grouped by refinement strategy (Similarity Threshold $\tau$ vs. Top-N Selection) with varying $k$.}
\label{tab:retrieval_refinement}
\end{table}

\begin{table}[t]
\centering
\resizebox{\columnwidth}{!}{%
\begin{tabular}{>{\centering\arraybackslash}m{3.2cm} >{\centering\arraybackslash}m{1.6cm} c c c c}
\toprule
\textbf{Model} & \textbf{KG} & \textbf{Hops} & \textbf{P} & \textbf{R} & \textbf{F1} \\
\midrule

\multirow{8}{=}{Qwen-2.5-7B-Instruct}
& \multirow{4}{=}{UMLS}
    & $k = 1$   & 0.3087 & 0.3098 & 0.2691 \\
&   & $k = 2$   & 0.1738 & 0.1757 & 0.1466 \\
&   & $k \leq 2$ & 0.1994 & 0.1364 & 0.1317 \\
&   & $k \leq 3$ & 0.1069 & 0.0604 & 0.0618 \\
\cmidrule(lr){2-6}

& \multirow{4}{=}{PrimeKG}
    & $k = 1$   & 0.2137 & 0.2995 & 0.1995 \\
&   & $k = 2$   & 0.0801 & 0.1592 & 0.0935 \\
&   & $k \leq 2$ & 0.0362 & 0.1178 & 0.0505 \\
&   & $k \leq 3$ & 0.0619 & 0.0730 & 0.0515 \\
\midrule

\multirow{8}{=}{Llama-3.1-8B-Instruct}
& \multirow{4}{=}{UMLS}
    & $k = 1$   & 0.2174 & 0.3650 & 0.2420 \\
&   & $k = 2$   & 0.1708 & 0.2746 & 0.1705 \\
&   & $k \leq 2$ & 0.1377 & 0.2556 & 0.1488 \\
&   & $k \leq 3$ & 0.0686 & 0.1112 & 0.0768 \\
\cmidrule(lr){2-6}

& \multirow{4}{=}{PrimeKG}
    & $k = 1$   & 0.1735 & 0.3451 & 0.1926 \\
&   & $k = 2$   & 0.0356 & 0.0974 & 0.0477 \\
&   & $k \leq 2$ & 0.0289 & 0.0962 & 0.0387 \\
&   & $k \leq 3$ & 0.0369 & 0.1216 & 0.0491 \\
\bottomrule
\end{tabular}%
}
\caption{Performance of KG-augmented supervised fine-tuning on UMLS and PrimeKG. We compare Llama-3.1-8B-Instruct and Qwen-2.5-7B-Instruct models trained with specific ($k=1,2$) versus cumulative ($k \le 2, 3$) retrieval contexts.}
\label{tab:logoskg_sft}
\end{table}

\subsection{Evaluation of Refinement Techniques for Filtering in Round 1}
\label{appendix:refinement}
We also optimized the retrieval results to reduce noise and examined whether refinement could improve performance. Following a setup common in prior work (e.g., Dr.Knows \citep{gao2025leveraging} and GraphRAG \citep{han2501retrieval}), we used SapBERT to compute the maximum cosine similarity between candidate diagnoses and query entities. All results in Table \ref{tab:retrieval_refinement} were computed using GPT-5-mini with ProbSum. To strictly investigate the impact of ranking and pruning on initial retrieval, these experiments utilize only LogosKG's Round 1 filtering output, excluding the Round 2 expansion. We explored two refinement strategies: a threshold-based method ($\tau = 0.7, 0.8, 0.9$) and a Top-N selection method ($N = 10, 15, 20, 25, 30$). 

We found that the threshold method worked well for reducing noise on UMLS, but it struggled with PrimeKG. It appears that the performance drop for PrimeKG was due to the method being too aggressive, likely filtering out useful evidence. On the other hand, the Top-N strategy proved to be much more reliable. By keeping a fixed number of the best candidates instead of relying solely on a strict score, it struck a better balance between filtering out noise and retaining key information. This approach led to consistent gains in Precision and F1 score, particularly in later rounds, where error management is important.

\subsection{\textsc{LogosKG}-Augmented Supervised Fine-Tuning}
\label{appendix:logoskg_sft}
We also investigated whether \textsc{LogosKG} could enhance training for downstream medical diagnosis. We approached this by retrieving evidence in two distinct modes: ``at specific hops'' (isolating entities at exact distances) and ``within specific hops'' (aggregating all entities up to a certain distance). We then augmented the original training data by appending these retrieved entities to the patient progress notes. Using this enriched input, we performed Supervised Fine-Tuning to teach the models to identify the gold-standard diagnosis within the provided context.

Our results in Table \ref{tab:logoskg_sft} clearly show that restricting retrieval to specific hops ($k=1, 2$) consistently outperforms cumulative strategies ($k \le 2, 3$). While cumulative retrieval captures more information, in practice, it overwhelms the model with noise. For example, for each sample in ProbSum, the number of diagnosis entities retrieved from PrimeKG increases drastically with depth, averaging 215 within 2 hops and rising to 42,533 within 5 hops. This noise actually drags performance below the baseline of standard supervised fine-tuning, which has the advantage of a cleaner and more focused search space. These findings underscore that effective KG-LLM training is not trivial. As \citet{khatwani2025brittleness} pointed out, simple fine-tuning is insufficient. Instead, the field is moving toward iterative, multi-agent frameworks, an exciting direction recently explored by \citet{zhao2025agentigraph,zuo2025kg4diagnosis,xie2025kerap}. We consider this as a key area for future work, where \textsc{LogosKG} can serve as the efficient retrieval backbone necessary for such advanced systems.

\end{document}